\documentclass[letterpaper, 10 pt, conference]{ieeeconf}
\IEEEoverridecommandlockouts
\let\labelindent\relax
\usepackage{enumitem}
\usepackage{balance}
\usepackage[font={footnotesize}]{caption}
\usepackage{subcaption}
\usepackage{array}
\usepackage{textcomp}
\usepackage{mathtools, nccmath}
\usepackage{graphicx}
\usepackage{amsfonts}
\usepackage{amsmath}
\usepackage{amssymb}
\usepackage{algorithm}
\usepackage{algorithmicx}
\usepackage{tikz}
\usepackage{arydshln}
\usepackage{multirow}
\usepackage{bm}
\usepackage{epstopdf}
\usepackage{siunitx}
\usepackage{xcolor}
\usepackage{tabularx}
\usepackage{cite}
\usepackage{listings}
\usepackage{algpseudocode}
\usepackage{bbold}
\usepackage{hyperref}

\usepackage{booktabs}       
\makeatletter
\let\NAT@parse\undefined
\makeatother
\newcommand{\figref}[1]{Fig.~\ref{#1}}

\newcommand{\stateName}{LLM-State }

\definecolor{codegreen}{rgb}{0,0.6,0}
\definecolor{codegray}{rgb}{0.5,0.5,0.5}
\definecolor{codepurple}{rgb}{0.58,0,0.82}
\definecolor{backcolour}{rgb}{0.95,0.95,0.92}

\lstdefinestyle{mystyle}{
    backgroundcolor=\color{backcolour},   
    commentstyle=\color{codegreen},
    keywordstyle=\color{magenta},
    numberstyle=\tiny\color{codegray},
    stringstyle=\color{codepurple},
    basicstyle=\ttfamily\footnotesize,
    breakatwhitespace=false,         
    breaklines=true,                 
    captionpos=b,                    
    keepspaces=true,                 
    numbers=left,                    
    numbersep=5pt,                  
    showspaces=false,                
    showstringspaces=false,
    showtabs=false,                  
    tabsize=2
}

\title{\LARGE \bf LLM-State: Open World State Representation for Long-horizon Task Planning with Large Language Model}

\author{
Siwei~Chen$^{1}$, 
Anxing~Xiao$^{1}$, 
David Hsu$^{1,2}$
\thanks{$^1$ School of Computing, National University of Singapore, Singapore.
\tt\small\{siwei-15, anxingx, dyhsu\}@comp.nus.edu.sg.
}
\thanks{$^2$ Smart Systems Institute, National University~of Singapore, Singapore.}
}

\begin{document}
\maketitle
\begin{abstract}

This work addresses the problem of long-horizon task planning with the Large Language Model (LLM) in an open-world household environment. Existing works fail to explicitly track key objects and attributes, leading to erroneous decisions in long-horizon tasks, or rely on highly engineered state features and feedback, which is not generalizable.
We propose an open state representation that provides continuous expansion and updating of object attributes from the LLM's inherent capabilities for context understanding and historical action reasoning. Our proposed representation maintains a comprehensive record of an object's attributes and changes, enabling robust retrospective summary of the sequence of actions leading to the current state. This allows continuously updating world model to enhance context understanding for decision-making in task planning. We validate our model through experiments across simulated and real-world task planning scenarios, demonstrating significant improvements over baseline methods in a variety of tasks requiring long-horizon state tracking and reasoning. (Video\footnote{Video demonstration: \url{https://youtu.be/QkN-8pxV3Mo}.})

\end{abstract}

\section{Introduction}
\label{sec:introduction}

Robot task planning in open-world household environments is critical as it enables efficient automation of daily chores, enhances the quality of life for residents, and promotes accessibility for individuals with physical limitations.
However, the open-world household environment poses a significant challenge in robot task planning, entrenched in factors such as unknown transition models, diverse goals, and arbitrary objects with limited predefined action types. Current Large Language Models' (LLMs) advanced abilities, like common-sense reasoning, enable them to tackle these challenges~\cite{vemprala2023chatgpt, openai2023gpt4}. Despite not requiring the formal definition of state space and transition models, these LLM-based methods either utilize a massive array of ground truth objects or manually engineered state representation, proving insufficient for long-horizon tasks in open-world settings~\cite{huang2022language,huang2023inner,singh2023progprompt,song2023llmplanner,rana2023sayplan}. Amplified by the partial observability, it becomes more daunting for LLMs to manage a flood of textual inputs to usefully track and analyze the unobserved but critical object attributes. Thus, discovering a new representation capable of enhanced object attribute tracking coupled with comprehensive context understanding is a pressing need in the field. This would improve decision-making in LLM's critical task planning tasks, leading to well-adapted systems for new tasks and environments.

\begin{figure}
    \centering
    \includegraphics[width=8.5cm]{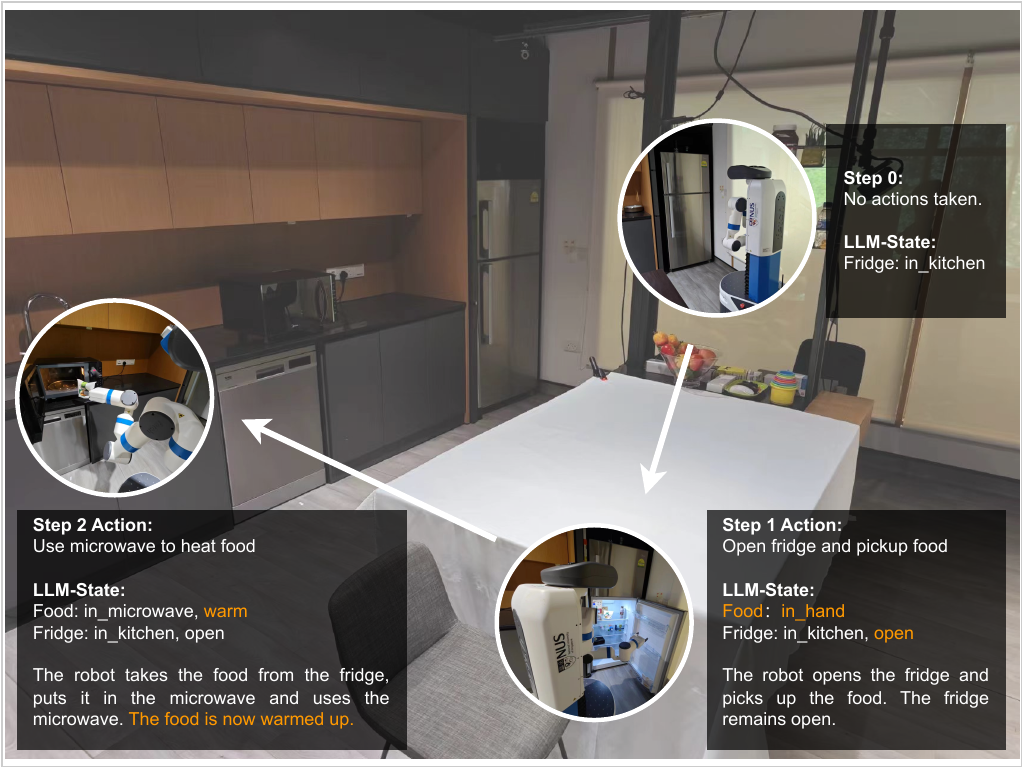}
    \caption{
    \stateName Example. The proposed state representation is a mixture representation of both structured objects with attributes that are automatically expanded and tracked by LLM, and an unstructured summary of historical data. For instance, when the robot takes food from the fridge (step 1) and uses a microwave to heat it (step 2), our \stateName auto-expands and tracks the unseen object key attributes through logic-based reasoning.  
    }
    \label{fig:main_example}
    \vspace{-0.75cm}
\end{figure}

In this work, we propose a new state representation for open-world task planning with Large Language Models (LLMs).
This representation enables updating a world model through a trial-and-error approach, leading to a more accurate estimation of the current state, action preconditions, and unknown transition effects.
Our approach combines a structured object-centric representation with an unstructured summary of historical actions and observed failures. The structured object-centric representation describes the world as a list of objects, each with different attributes indicating their current status. For example in \figref{fig:main_example} task \textit{heat up some food and pass it to the user}, the state could be represented as \textit{Food: in microwave, warm}, where \textit{in microwave} and \textit{warm} are the key attributes of the food item. To construct this structured representation, we first employ the \textit{LLM as Attention} to identify relevant objects from a long list of image object detection results. Then, we use the \textit{LLM as State Encoder} with the identified objects, robot-executed actions as inputs, and new object attributes as outputs. For example, LLM takes in “the robot put the food in the microwave and switch on”  and outputs state \textit{Food: in microwave, warm}. 
This object-centric representation is open in the sense that the number of objects and their attributes can be expanded and updated dynamically, providing a more flexible representation for \textit{LLM as Policy} to generate actionable commands for the robot.

However, relying solely on a structured object-centric representation is not sufficient for long-horizon task planning in open-world environments. There are two main challenges: (1) as the task horizon increases, the predicted object attributes may become less accurate due to large amounts of input information, and (2) some critical information may be missing from the structured representation, such as failed robot actions and their underlying causes. To address these challenges, we propose an additional unstructured representation that first summarizes the robot's action history through a step-by-step chain of thought processes. This summary enables the LLM to predict object attributes more accurately via a step-wise breakdown and chain-of-thought process. Furthermore, the unstructured representation pays special attention to recent failed robot actions, and reasons about the possible causes of the failure. One type of common failure is unmet action preconditions, such as placing items in a cabinet without opening it first. Such failure reasoning can be captured in our unstructured summary and used to inform the LLM's policy for generating new actions for complex, open-world tasks.

In summary, we propose a novel state representation for task planning with LLM in the open-world, constructed and updated automatically by LLM for object state tracking and reasoning. This representation accommodates new object attributes and incorporates retrospective summary for better attribute prediction and failure recovery, significantly improving long-horizon task planning performance in open-world scenarios. We assume a perfect image-based object detector and robust low-level action controller.

\section{Related Work}
\label{sec:related}

\subsection{Task Planning with LLMs} Recent advances in LLMs demonstrate impressive reasoning capabilities for high-level decision making in robotic tasks. A variety of studies explore how to leverage the common sense knowledge from LLMs to solve complex planning tasks, by decomposing instructions into action sequences~\cite{brohan2023can,huang2023inner, huang2022language, valmeekam2022large, silver2022pddl, song2023llmplanner, driess2023palm, liu2023reflect, wu2023tidybot, rana2023sayplan}, generating executable code~\cite{singh2023progprompt, liang2023code, vemprala2023chatgpt, silver2023generalized}, translating natural language into formal specifications \cite{xie2023translating, skreta2023errors, liu2023llm+}, and providing auxiliary information for classical planners\cite{zhang2023large,ding2023task, ding2023integrating, zhao2023large}.
Most works assume the availability of all objects and their accurate attributes in the environment \cite{brohan2023can,singh2023progprompt,rana2023sayplan,zhao2023large, silver2023generalized} or directly detect them from the perception system \cite{huang2023inner,liang2023code,ding2023integrating,liu2023reflect,wu2023tidybot,song2023llmplanner}, using them as state representations for LLMs to reason and plan. However, these predefined state representations may not capture the complexity of certain real-world domains. Some works employ self-reflection techniques \cite{wang2023describe,shinn2023reflexion} or external models for failure explanation\cite{liu2023reflect}, but these free-form summaries may be inefficient for complex, long-horizon tasks requiring explicit state tracking.
Related research also investigates accepting encoded images as state representations and outputting subsequent actions \cite{driess2023palm,brohan2023rt,lin2023learning}. 
where the perception and reasoning modules are coupled. However, these models typically require large amounts of domain-specific multi-modal data and may lack generalization across diverse domains. To achieve flexibility where not losing generalizability, our work relies on the perception-reasoning decomposition, and as an interface between perception and reasoning, our state representation strikes a balance between structure and flexibility, which improves efficiency and generalization.

\subsection{State Representations in Task Planning} Task planning, or symbolic planning, involves deciding on action sequences to achieve goals based on initial state and action effects descriptions. State representations are a key aspect of task planning, as they define the landscape in which planning occurs. In classical planning \cite{fikes1971strips, fox2003pddl2, brewka2011answer}, these representations are predefined, often using predicates to track state changes. LLM-based planning systems can adapt to various problem descriptions and avoid the need to define all predicates and transition models. Some approaches use exhaustive object and attribute descriptions in prompts \cite{ silver2022pddl, valmeekam2022large, singh2023progprompt}, while others rely on hand-engineered and task-specific state representations \cite{huang2022language, liang2023code,wang2023describe, song2023llmplanner, rana2023sayplan}. Recent methods explore dynamic memory with reflection \cite{shinn2023reflexion}, semantic search in scene graph\cite{rana2023sayplan}, multi-sensory summary \cite{liu2023reflect}, and training task-conditioned state descriptions \cite{nottingham2023selective} to achieve more flexible state representations. However, a major challenge in existing LLM-based planning systems is their inability to explicitly track flexible state changes, particularly those involving undefined predicates. This limitation affects both consistency and efficiency in open-world planning. 
In this work, we explore the action reasoning ability of LLM to automatically and explicitly construct and track the structured objects with comprehensive context understanding, enabling LLM to solve long-horizon tasks effectively.

\section{Problem Formulation}
\label{sec:problem_formulation}

In this work, we address the problem of open-world robot task planning in unseen home environments, where a robot must perform tasks according to the open natural language instructions. 
\subsection{Input and Output}
\label{sec:input_ouput}
The system takes the following inputs: \textbf{Goal Instruction in Text ($I$)}; \textbf{Step-wise Observation} $(\mathcal{O}_t, \mathcal{P}_t, \mathcal{H}_t, l_t)$, where $\mathcal{O}_t$, $\mathcal{P}_t$, $\mathcal{H}_t$, and $l_t$ indicate the name of observed objects, rooms, holding objects, and the robot's current location, respectively;  
\textbf{Pre-defined Primitive Actions ($A$)}, including move($o$), pickup($o$), placein($o_1, o_2$), placeon($o_1, o_2$), open($o$), close($o$), switchon($o$), switchoff($o$) and wait(), where $o$ and $c$ are the name of objects and containers. The action types are predefined but the object arguments are open, e.g. without predefined objects in the domain; \textbf{Primitive Actions Binary Result ($r_t$):} Binary number indicating success or failure results for each primitive action at time step $t$.

The system outputs \textbf{Step-wise Primitive Action Commands ($a_t \in A$):} A sequence of primitive actions with object argument to accomplish the given goal. Note that our system works in a closed-loop fashion, so it may need to go through multiple planning processes to solve one task.

\begin{figure*}[t]
    \centering
    \includegraphics[width=0.98\textwidth]{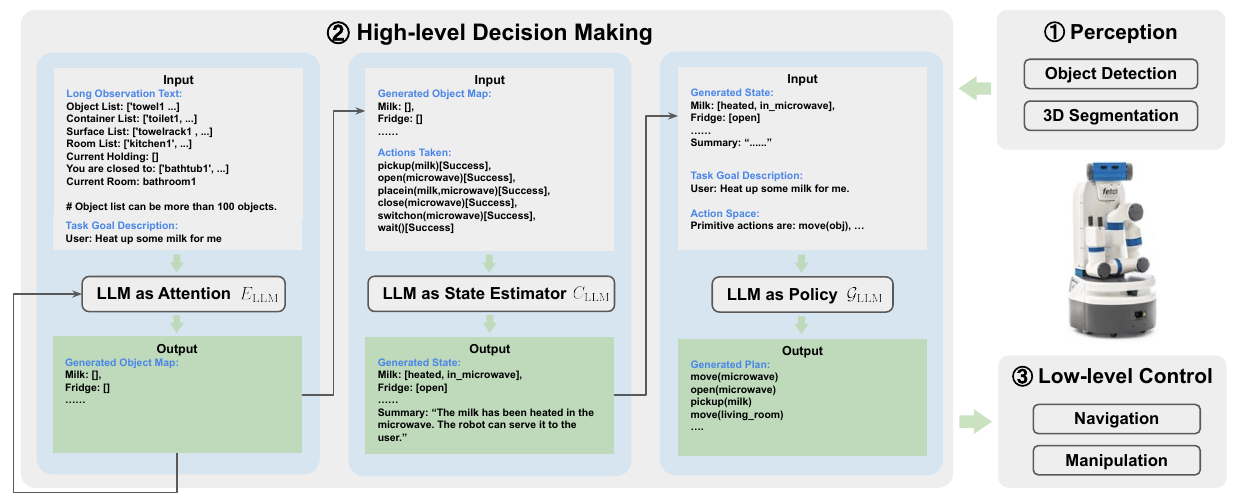}
    \caption{
Overview of the system framework. The task planner consists of three components: LLM as Encoder, LLM as State Estimator, and LLM as Policy. The perception system output the observation to Task Planner. In the task planner, LLM encodes and tracks the \stateName representation, which will be used to assist plan generation. The generated action will be executed by the low-level controller.  }
    \label{fig_framework}
    \vspace{-0.5cm}
\end{figure*}

\subsection{Problem Definition}
We address the problem of open-world robot task planning. In our setting, the open world represents the planning system don't have full access to the underlying planning domain, e.g. complete object list with ground-truth states, and accurate action precondition and effect. Only the observation $(\mathcal{O}_t, \mathcal{P}_t, \mathcal{H}_t, l_t)$  and Primitive Actions ($A$) are available with the format of the name, leading to unconstrained state space and unknowing transition.

The objective is to generate and execute step-wise primitive action commands $a_t$ to effectively accomplish the goal given $I$, $A$, $(\mathcal{O}_t,  \mathcal{P}_t, \mathcal{H}_t, l_t)$, and $R_t$. The problem assumes a perfect object detector to map the image observation to the object list in $\mathcal{O}_t$, reliable action execution with binary success indicator, and navigable environments. 
The core idea behind our method is to construct the state representation $s_t$ based on the historical information $I$, $A_t$, $(\mathcal{O}_t,  \mathcal{P}_t, \mathcal{H}_t, l_t)$ to support LLM planner in open-world robot task planning. 
The method performance will be evaluated on household tasks both in simulation and Real-world robots.

\section{Method}

\subsection{Overview}
The heart of our proposed method to address the open-world planning challenges lies in its state representation. This unique representation is a mixture of structured object-centric
representation and an unstructured summary of historical actions with observed failures analysis.
\figref{fig_framework} demonstrates the key processes of our system. 
The perception system with object detector and segmentation model outputs the observation $\mathcal{O}_t$ to our task planner. In the task planner, LLM as Attention and LLM as State Estimator encodes and tracks the \stateName representation, which will be used to assist plan generation using LLM as Policy. The generated action will be executed by the low-level controller to accomplish the navigation and manipulation tasks. 



\begin{algorithm}[t]
\small
    \caption{Task Planning Execution Flow}
    \begin{algorithmic}[1]
        \Require  I: Goal Instruction, A: Primitive Actions \par
           \quad O: Observed Objects, $E_{\text{LLM}}$: LLM as Attention\par
           \quad $C_{\text{LLM}}$: LLM as State Estimator,  $\mathcal{G}_{\text{LLM}}$: LLM as Policy
    \State Initialize history action list $A_h \gets []$
    \State Initialize key objects $\hat{s}_0 \gets []$
    \While{not done}
    \State Update key objects $\hat{s}_t \gets E_{\text{LLM}}(O_t, I) + \hat{s}_{t-1}$
    \State Update state representation $s_t \gets C_{\text{LLM}}(\hat{s}_t, A_h, O_t)$
    \State Generate $n$ actions $a_{t:t+n} \gets \mathcal{G}_{\text{LLM}}(s_t, O_t, I)$
    \For{$a_t$ in $a_{t:t+n}$}
    
    \State $O_t, r_t \gets \text{execute}(a_t) $ \# $r_t$: binary success indicator
    \State Append actions, $A_h \gets A_h + (a_t, r_t)$
    \If{not $r_t$}
    \State break \# action failed, re-plan
    \EndIf
    \EndFor
    \EndWhile

    \end{algorithmic}

    \label{exe_flow}
    \vspace{-0cm}
\end{algorithm}

The details of the execution flow for task planning can be found in Algorithm \ref{exe_flow}, our system consists of three core functions: LLM as Attention $E_{\text{LLM}}$, LLM as State Estimator $C_{\text{LLM}}$, and LLM as Policy $G_{\text{LLM}}$. 
The LLM as Attention $E_{\text{LLM}}$ will extend the list of key objects $\hat{s}_t$ relevant to the task $I$ based on the latest observed object list $\mathcal{O}_t$. LLM as State Estimator $C_{\text{LLM}}$ uses the key objects list $\hat{s}_t$, historically executed action $A_h$ with binary success indicator $R_t$, and observation $\mathcal{O}_t$ to generate the state representation $s_t$ with state change for the object in the key objects list and the summary. The LLM as Policy function $G_{\text{LLM}}$, generates the action sequences based on the current state representation $s_t$, observation $\mathcal{O}_t$ and instruction $I$. The generated action in code format is automatically executed by the robot to interact with the environment. The state will be updated after an action fails and when all planned actions are executed successfully.

\subsection{\stateName Representation}
\label{state_representation}
\begin{figure}[h]
    \centering
    \includegraphics[width=7cm]{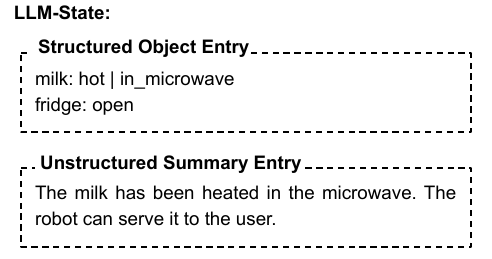}
    \caption{
    \stateName Example of \stateName Representation. 
    }
    \label{fig:LLM_state_example0}
\end{figure}

The proposed large language model's (LLM) state representation operates as a task-specific, expandable dictionary. As shown in \figref{fig:LLM_state_example0}, this representation, built on key-value pairs, consists of many object entries with different attributes, and a retrospective summary entry that reasons about the history context especially for the action failures. We discuss the details of each entry type below.

\subsubsection{Structured Object Entry.}
Structured Object Entry consists of a dictionary where each object key represents an object relevant to the task goal. The key's value contains a list of descriptive attributes, such as status, location, and condition. These objects and their attributes are dynamically constructed and updated by the LLM based on observations and reasoning.
For example, when a robot prepares breakfast, it retrieves milk, places it in the microwave, and heats it. The milk becomes hot, an attribute that's not easily observable but can be logically deduced. Failing to update and track this could cause errors in long-horizon tasks, such as reheating already hot milk. This state representation addresses open-world environment challenges, enabling efficient tracking, recognition, and understanding of new objects and attributes for task planning.

\subsubsection{Unstructured Retrospective Summary Entry.}

We propose an additional unstructured object representation that first summarizes the robot's action history in a step-by-step manner, see \figref{fig:LLM_state_example}. This process serves as a ``Chain of Thoughts" that helps predict better object attributes by following a sequential thought process. One detailed example can be found in Appendix \ref{sec:summary_examples}, where using the unstructured summary can better predict the object attributes such as the cabinet1 is closed.
By maintaining a coherent narrative of the robot's actions, this representation aids in understanding the current state.

Secondly, the unstructured object representation focuses on more valuable information, such as recent failed robot actions. It reasons about the possible causes of the robot action failure and provides insightful suggestions to recover from the failure. 
In our setting, potential causes of failure can arise from unknown preconditions (e.g., not realizing that the hand is already occupied, or performing the same opening action repeatedly) and unexpected transitions (e.g., trying to pick up an object that can't be picked up, or trying to open a container that is locked).

These suggestions are highly beneficial and can bias the action selection when using LLM as a policy in later stages. For instance, consider the example in \figref{fig:LLM_state_example}, where the robot fails to put breadslice2 into the toaster. The LLM generates possible causes for this failure, such as the presence of another breadslice already inside the toaster, which prevents it from accommodating more bread slices simultaneously. By identifying the cause of the failure, the unstructured retrospective summary can guide the robot toward a suitable recovery strategy, such as removing the existing bread slice before attempting to insert breadslice2.

In summary, the unstructured retrospective summary entry not only serves as a chain of thoughts for better object attribute prediction but also offers valuable insights into overcoming task failures and improving the overall performance of the robot.

\subsection{Construct and Use the State Representation}

\subsubsection{LLM as Attention $E_{\text{LLM}}$}
Directly feeding a large number of observed objects into state construction and planning with Large Language Model (LLM) can lead to a lack of focus on relevant objects. Furthermore, the long list of observations in the prompt will also exceed the LLM's context length limit.
To avoid the above issues
we use the LLM to extract the context-aware objects relevant to the task first.
As depicted in the left column of \figref{fig_framework} (with detailed prompts provided in Appendix \ref{appendix:prompt}), the LLM ingests the raw observation $O_t$ and a text-based instructional goal $I$. The LLM then generates output in the form of structured, executable code, as shown below:

\begin{figure}[t]
    \centering
    \includegraphics[width=8.5cm]{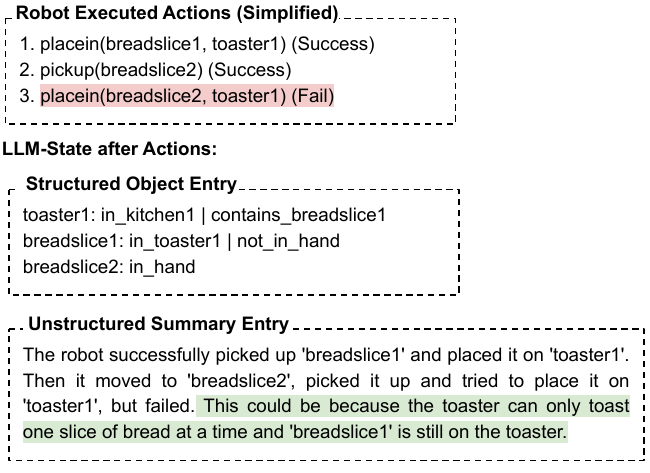}
    \caption{
    \stateName Example. After placing a slice of bread in the toaster, the robot fails to place another one (highlighted in red). The unstructured summary offers additional context about the failure (highlighted in green). 
    }
    \label{fig:LLM_state_example}
        \vspace{-0.6cm}
\end{figure}

\begin{lstlisting}[language=Python, caption=Function to include related objects]
    add_related_objects("object") 
\end{lstlisting}

Each function call adds the corresponding object into the relevant object name list $\hat{s}_t$. This process is recursive and designed as an ``only-add'' mechanism, ensuring the continuous expansion of the tracked object list without erasing previous entries. This state dictionary thus serves as a representation of the attention given to related objects, embedded within the state representation itself. 

After executing the code, the state dictionary will be updated to $\hat{s}_t$, a name list of all the relevant objects. We use $E_{\text{LLM}}$ to represent the process of LLM as Attention. In this context, $E_{\text{LLM}}(O_t, I)$ represents the attention mechanism of the LLM, where $O_t$ is the current observation and $I$ is the instructional goal. The attention mechanism determines which objects in the observation are relevant to the goal and should be added to the state representation. Therefore, we have 
$$\hat{s}_t \gets E_{\text{LLM}}(O_t, I) + \hat{s}_{t-1},$$
which means the current object list is updated by adding the objects identified by the attention mechanism to the previous relevant object name list.

\subsubsection{LLM as State Estimator $C_{\text{LLM}}$}
Our large language model (LLM) utilizes its integrated reasoning ability to manage updates to objects' state attributes,
$$s_t \gets C_{\text{LLM}}(\hat{s}_t, A_h, O_t),$$
using state dictionary $\hat{s}_t$ provided by LLM as Attention, and a history of executed actions and their outcomes (success or failure) $A_h$. Detailed prompt can be found in Appendix \ref{appendix:prompt}.
This approach utilizes all previous actions from step 0 onward, preventing error accumulation. Inaccuracies in LLM may cause incorrect conclusions in recursive settings; however, by estimating from the beginning, errors can be resolved without carrying over to subsequent state estimations, improving overall accuracy.

The LLM generates two types of commands that achieves the state estimation process:

\begin{lstlisting}[language=Python, caption=Function to update state representation ]
    generate_summary("Fill in the reasoning here, especially for the possible failure cases.")

    add_attribute("object", "attr_1 | attr_2")
\end{lstlisting}

In the context of the first command, "generate\_summary()", the LLM interprets and reiterates the current scenario, applying its contextual understanding to try and identify the potential reasons behind any action failures. 
This approach is advantageous for two reasons: (a) It distills a potentially verbose history of observations and actions into a concise and understandable summary of history; (b) It encourages the identification and understanding of the causes behind any action failures, allowing the LLM to inform the downstream policy better, potentially enabling it to avoid similar mistakes.

In the second function, "add\_attribute()", the LLM reviews previous actions and their context, and it extrapolates the logical implications of these actions on an object's attributes. For instance, if a robot moves milk from the fridge to a microwave and heats it, the milk becomes hot.  This change is not directly observable but can be logically inferred. Hence, our proposed state representation addresses the challenges posted by open-world environments, like tracking, identifying, and understanding new objects and new attributes, thus enhancing task planning significantly. After executing the code, the state will be updated to $s_t$. We use $C$ to represent the process of LLM as State Estimator, then we have $s_t \gets C(\hat{s}_t, A_h, O_t)$.

\begin{table*}
\caption{Results in Simulation for four methods: InnerMonologue-Windowed (InnerM-W), ProgPrompt, Baseline method, and  (Ours). An asterisk (*) highlights unique challenges within certain tasks, such as the inability to grasp specific object types. The results are evaluated according to the Success Rate (SR) and Average Steps (AS) required. }

\label{table:simulation_result}

\begin{tabular}{@{}llllllllll@{}}
\toprule
                              &                                                                                                                    & \multicolumn{2}{c}{InnerM-W} & \multicolumn{2}{c}{ProgPrompt} & \multicolumn{2}{c}{Our w/o States} & \multicolumn{2}{c}{Ours} \\ \cmidrule(lr){3-4} \cmidrule(lr){5-6} \cmidrule(lr){7-8} \cmidrule(lr){9-10}
                              & Tasks                                                                                                              & SR           & AS          & SR            & AS           & SR             & AS            & SR          & AS         \\ \midrule
\multirow{8}{*}{Simple} &``Switch off the light in bedroom."  & 5/5 & 4.0 & 5/5 & 7.0 & 5/5 & 6.2 & 5/5 & 6.2 \\
 & ``Open the fridge."   & 5/5 & 4.2 & 5/5 & 3.0 & 5/5 & 4.0 & 5/5 & 4.4 \\
 & ``Put a book on the desk."  & 3/5 & 19.4 & 5/5 & 5.2 & 5/5 & 4.8 & 5/5 & 5.0 \\
 &``Switch off all lights in the house." & 1/5 & 29.6 & 5/5 & 15.0 & 3/5 & 24.4 & 5/5 & 18.8 \\
 & ``Use computer."  & 4/5 & 8.8 & 4/5 & 9.4 & 4/5 & 8.4 & 4/5 & 8.4 \\
 & ``Find a magazine, put it on a bed."  & 4/5 & 13.4 & 5/5 & 15.0& 3/5 & 16.6 & 4/5 & 14.8 \\
 & ``Make toast."  & 4/5 & 13.4 & 5/5 & 12.4 & 5/5 & 8.0 & 5/5 & 8.6 \\
 & ``Heat milk with microwave."  & 0/5 & 30.0 & 0/5 & 30.0 & 4/5 & 14.6 & 4/5 & 17.6 \\ 
 \midrule

& \textbf{Mean}&  \multicolumn{2}{c}{65.00\%} & \multicolumn{2}{c}{85.00\%} & \multicolumn{2}{c}{85.00\%} & \multicolumn{2}{c}{92.5\%} \\

\midrule

\multirow{7}{*}{Hard} & \begin{tabular}[c]{@{}l@{}}``Put toothbrush and face cream in the bathroom cabinet."\end{tabular} & 0/5 & 60.0 & 0/5 & 60.0 & 0/5 & 60.0 & 4/5 & 36.8 \\
 & \begin{tabular}[c]{@{}l@{}}``Empty the surface of kitchen table, and place them in kitchen cabinets."\end{tabular}   & 0/5 & 150.0 & 0/5 & 150.0 & 0/5 & 150.0 & 2/5 & 131.0 \\
 & \begin{tabular}[c]{@{}l@{}}``Get wine glass and juice, place them on desk."\end{tabular}   & 0/5 & 50.0 & 0/5 & 50.0 & 0/5 & 50.0 & 3/5 & 41.4 \\
 & \begin{tabular}[c]{@{}l@{}}``Make 3 slices of toasts using toaster and place them on kitchen table."\end{tabular}   & 0/5 & 60.0 & 0/5 & 60.0 & 2/5 & 55.8 & 5/5 & 40.0 \\
 & \begin{tabular}[c]{@{}l@{}}``Make 6 slices of toasts using toaster and place them on kitchen table."\end{tabular}   & 0/5 & 120.0 & 0/5 & 120.0 & 0/5 & 120.0 & 4/5 & 92.6 \\
 & \begin{tabular}[c]{@{}l@{}}``Make 2 slices of toasts using toaster and place them on kitchen table."*\end{tabular}   & 0/5 & 100.0 & 0/5 & 100.0 & 0/5 & 100.0 & 4/5 & 66.8 \\
 & \begin{tabular}[c]{@{}l@{}}``Take away three objects from the kitchen table."*\end{tabular}   & 0/5 & 100.0 & 0/5 & 100.0 & 1/5 & 86.8 & 5/5 & 38.2 
                              \\ 
                              \midrule
& \textbf{Mean} & \multicolumn{2}{c}{0.00\%} & \multicolumn{2}{c}{0.00\%} & \multicolumn{2}{c}{8.71\%}  & \multicolumn{2}{c}{77.14\%} \\
                              \bottomrule
\end{tabular}
\vspace{10pt}
\vspace{-0.8cm}

\end{table*}


\subsubsection{LLM as Policy $\mathcal{G}_{\text{LLM}}$}
We use the new representation to generate a plan. The current state $s_t$, current observations $O_t$, and task goal $I$ are given to the LLM. The output is a sequence of future primitive actions $a_{t:t+n}$ where $n$ is a  planning horizon. This generation process can be represented as:

\begin{equation}
a_{t:t+n} \gets \mathcal{G}_{\text{LLM}}(s_t, O_t, I)
\end{equation}

Where $\mathcal{G}$ denotes the LLM that has been prompted as a plan generator. The generated actions $a_{t:t+n}$ in text form will be executed in a way ``text as code" in the Python script to make corresponding function calls defined before in the problem formulation \ref{sec:problem_formulation}. Incorrectly formatted function calls from the LLM will be ignored and discarded.
Additionally, the state update can be triggered after an action fails or after every action. In Algorithm \ref{exe_flow} and experiment, we update the state after an action fails and when all planned actions are executed successfully.

\textbf{Exploration in Open-World.} For objects not present in the current state but required to complete a task, the LLM policy actively generates hypotheses using common sense. For example, it hypothesizes that milk could be located in the refrigerator and consequently, generates an action to open the refrigerator in search of milk. While such actions do not guarantee success, they consistently provide new observations that update our state and allow for re-planning based on the newly collected information.

\subsection{System and Implementation}
\label{subsection:system}
The overall proposed system developed in this article is illustrated in Fig. \ref{fig_framework}. This shows the integration and deployment of our method in a real-world robotics system.
Our system is a composite of three primary elements: the high-level decision-making module, the perception module, and the low-level controller. We use GPT-4 \footnote{The version of GPT-4 we used is gpt-4-0613} \cite{openai2023gpt4} to generae all the text responses from LLM. In the prompt, instead of ask LLM to generate free-form text response, we ask LLM to generate codes to call specific functions to construct, update the state representations and generate action plans.

\textbf{Perception.} The perception module, using RGB-Depth camera data, provides textual object detection results to the task planning and archives 3D coordinates for each object, enhancing navigation and manipulation. The robot covers a 360-degree radius by maneuvering to six distinct angles. Image tagging is obtained using the model in \cite{zhang2023recognize}, and Grounding DINO \cite{liu2023grounding} provides bounding boxes. Lastly, Segment-Anything \cite{kirillov2023segany} extracts masks, fused with RGB-D data for 3D frame back-projections.

\textbf{Low-level Controller.} The low-level controller consists of navigation and manipulation skills. For navigation, the robot utilizes occupancy grid maps with room names and positions and 3D coordinates of detected objects provided by perception modules to navigate using ROS built-in path and motion planning tools \cite{quigley2009ros}. For manipulation skills, an imitation learning policy is adopted, enabling complex actions like opening a microwave or grasping objects from a fridge. This involves the use of the VR controller, allowing the human demonstrator to control the robot's gripper with precision across six degrees of freedom and monitor its camera view remotely through screen.
The collected data is used to train models using Imitation Learning \cite{ross2011reduction}, taking RGB-D images and the robot arm's joint state as input and predicting joint angle movement sequences. Averaged 30 trajectories demonstrations are gathered per primitive action using the real robot.

\section{Experiments and Evaluation}
In this section, we evaluate the performance of our method using various open-world household tasks in both virtual simulation and a real-world with a mobile manipulator.

\begin{figure}[!h]
    \centering
    \includegraphics[width=8.5cm]{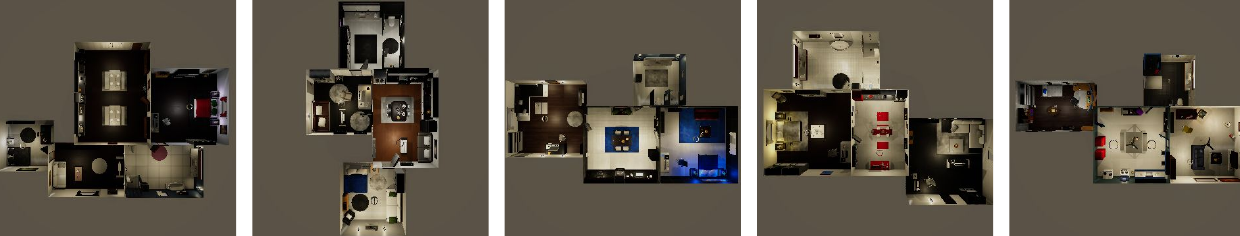}
    \caption{The five maps used in VirtualHome simulation.
    }
    \label{fig:house_map}
    \vspace{-0.6cm}
\end{figure}

\subsection{Simulation Experiments}

\subsubsection{Experimental settings in Simulation}
\label{subsubsection:simulation_setting}

We evaluate our method using VirtualHome \cite{puig2018virtualhome}, a large-scale household simulation with a variety of interactive objects, receptacles, and rooms. We focus on task planning performance and replace the perception module and low-level controller with simulation API calls. The observation and action space specifics can be viewed in the section \ref{sec:problem_formulation}. The robot can detect objects in the same room but not in closed receptacles. Experiments are conducted across five house maps shown in \figref{fig:house_map} and 15 tasks per map. Tasks are divided into Simple ( less than 30 steps) and Hard (more than 30 steps) levels for comprehensive evaluation.

We compare our method with three baselines: InnerMonologue~\cite{huang2023inner}, ProgPrompt~\cite{singh2023progprompt}, and a naive version of LLM without our \stateName representation. InnerMonologue often exceeds the maximum token limit for long-horizon tasks. To mitigate this, we maintain a window, referred to as InnerMonologue-W. This issue is common in methods that use long history observations and actions as state. ProgPrompt, generates code instead of free form text to perform task planning. For a fair comparison, we also extend ProgPrompt with a replanning procedure. For the naive LLM, we directly prompt the observations and action history to the LLM, similar to previous work \cite{song2023llmplanner}.

\begin{table}[]
\caption{Ablation Study in Simulation.}
\label{table:ablation_study}
\centering
\begin{tabular}{lllll}
\toprule
Task & \text{w/o States} & \text{w/o Summary} & \text{w/o Objects} & Ours \\
\midrule
Simple & 85.00\% & 92.50\% & 95.00\% & 92.5\% \\
Hard  & 8.71\% & 14.29\% & 64.86\% & 77.14\% \\

 \bottomrule
\end{tabular}
\vspace{-0.5cm}

\end{table}

\subsubsection{Simulation Experiments Results}

\begin{figure*}[tb]\label{fig: experiemnt results}
    \centering
    \begin{subfigure}[t]{0.195\linewidth}
        \centering
        \includegraphics[height = 2.6cm]{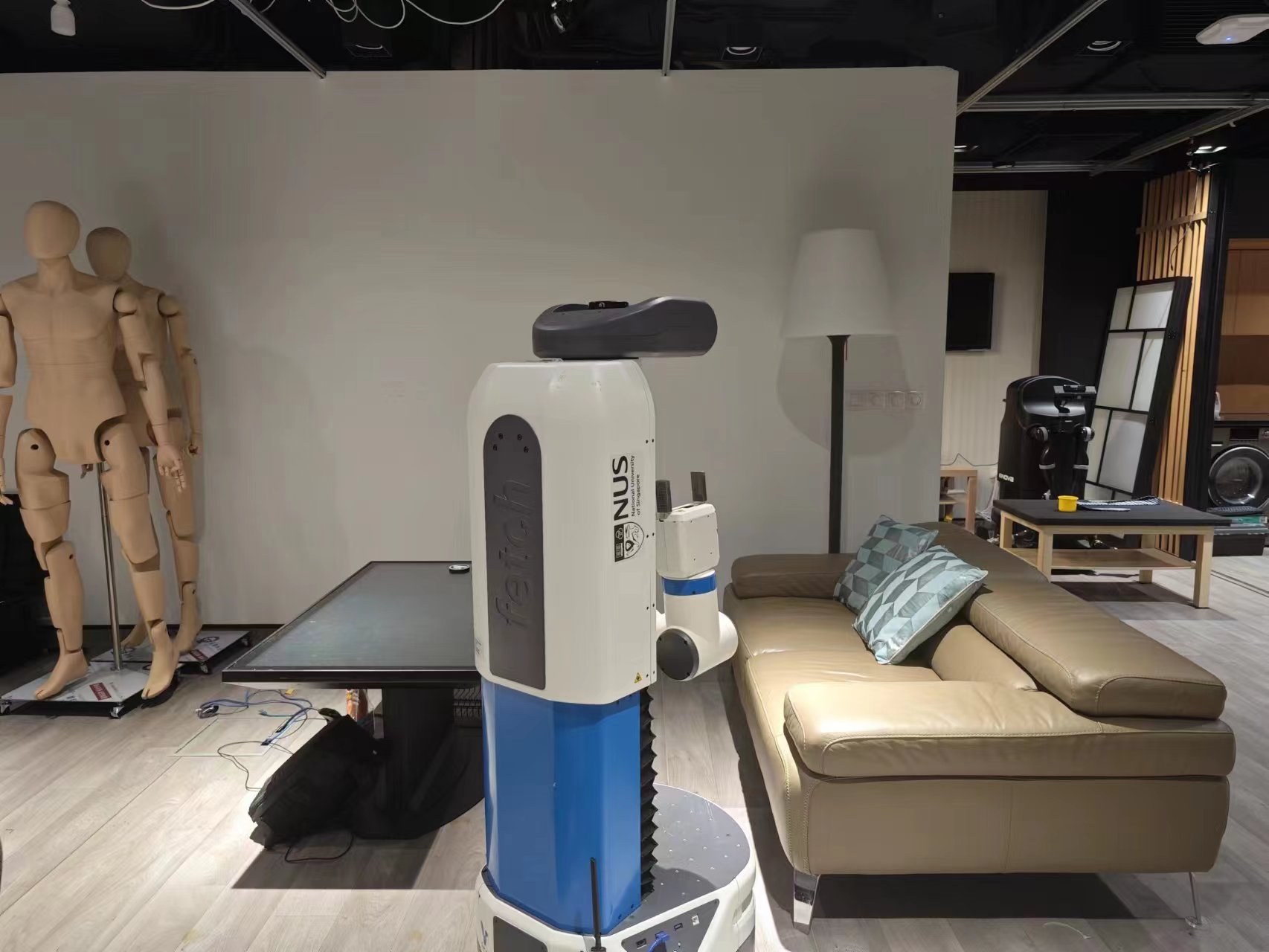}
        \caption{}
        \label{subfig:a}
    \end{subfigure}
    \begin{subfigure}[t]{0.195\linewidth}
        \centering
        \includegraphics[height = 2.6cm]{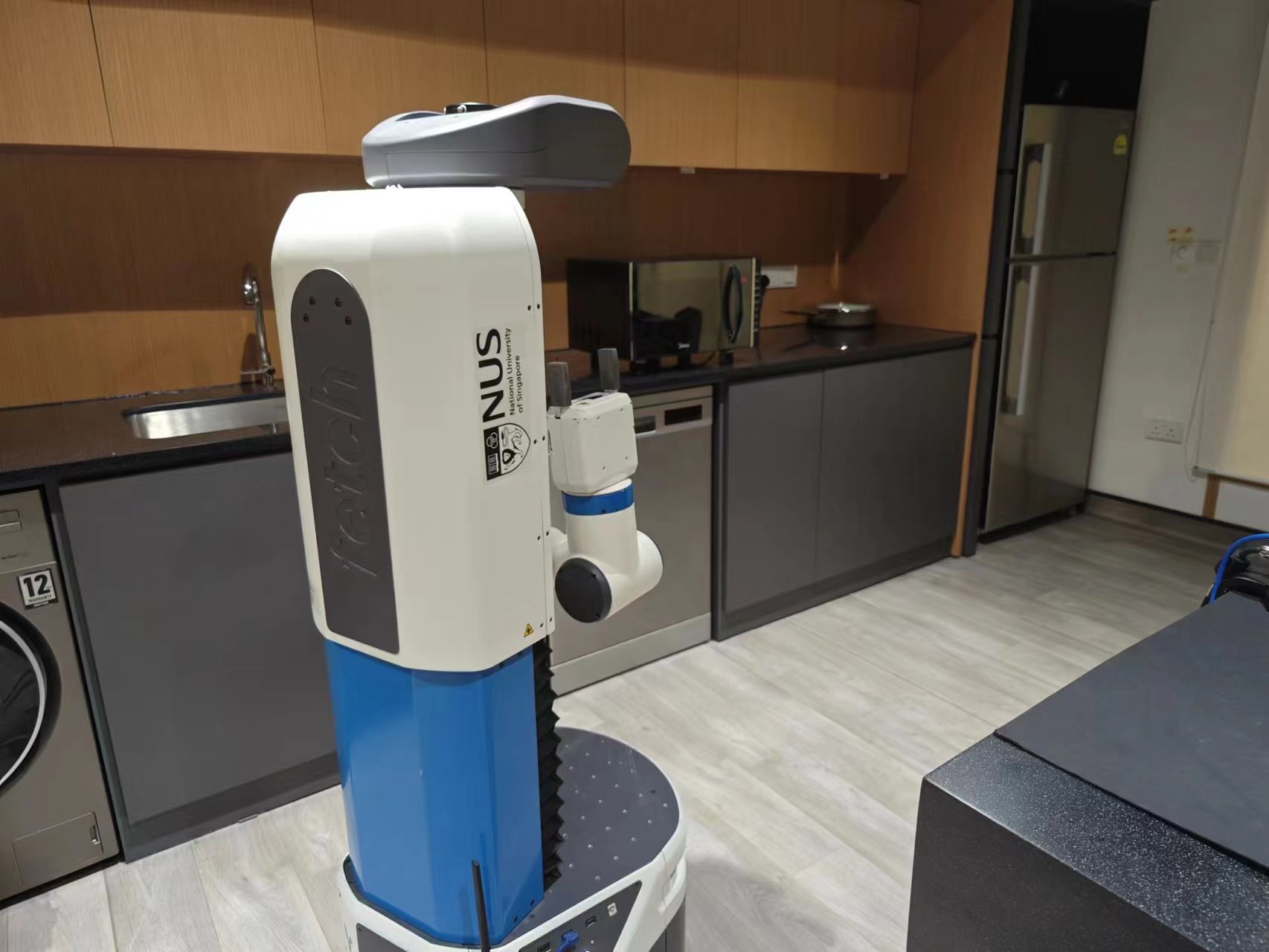}
        \caption{}
        \label{subfig:b}
    \end{subfigure} 
    \centering
    \begin{subfigure}[t]{0.195\linewidth}
        \centering
        \includegraphics[height = 2.6cm]{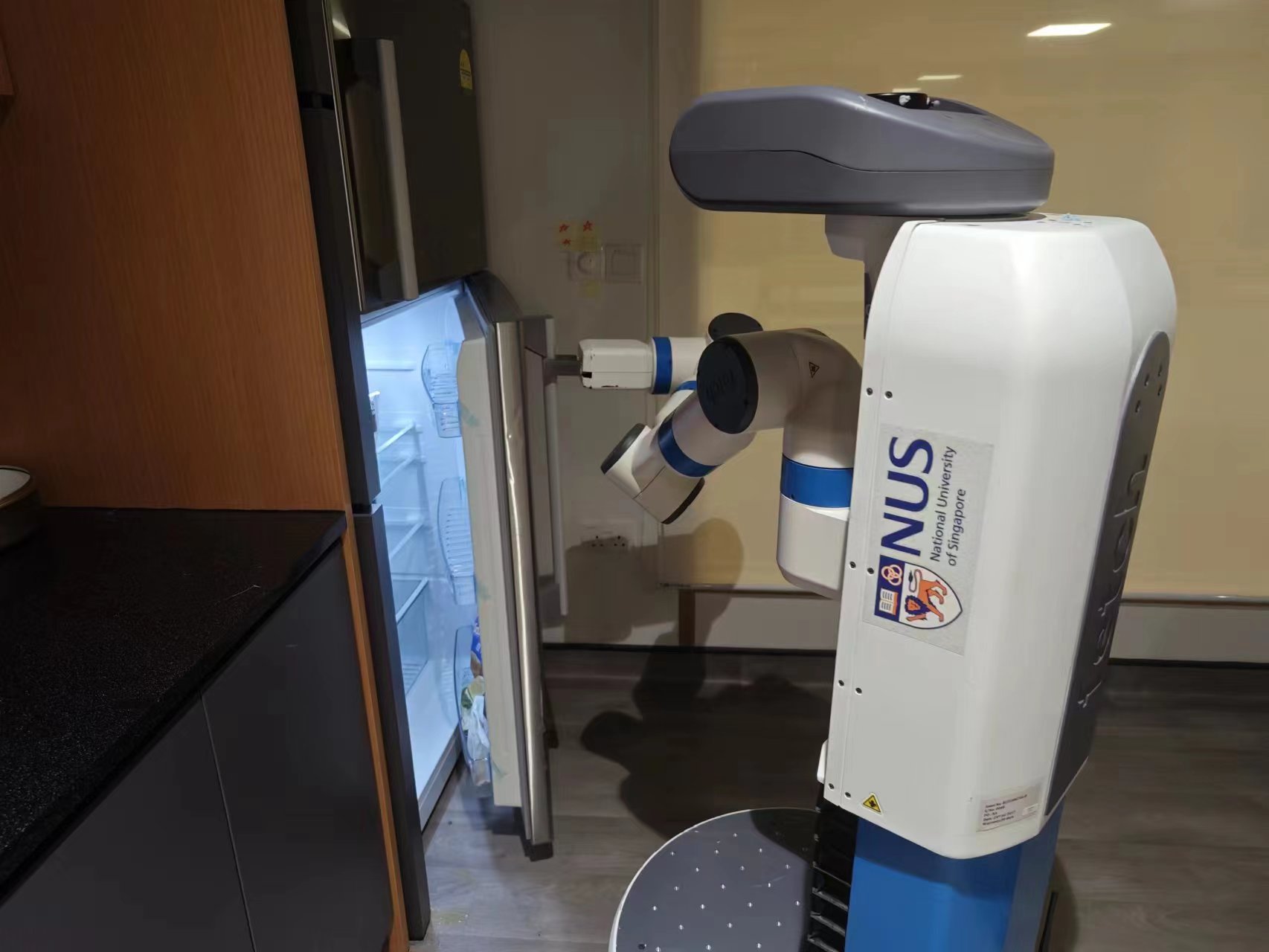}
        \caption{}
        \label{subfig:c}
    \end{subfigure}
    \begin{subfigure}[t]{0.195\linewidth}
        \centering
        \includegraphics[height = 2.6cm]{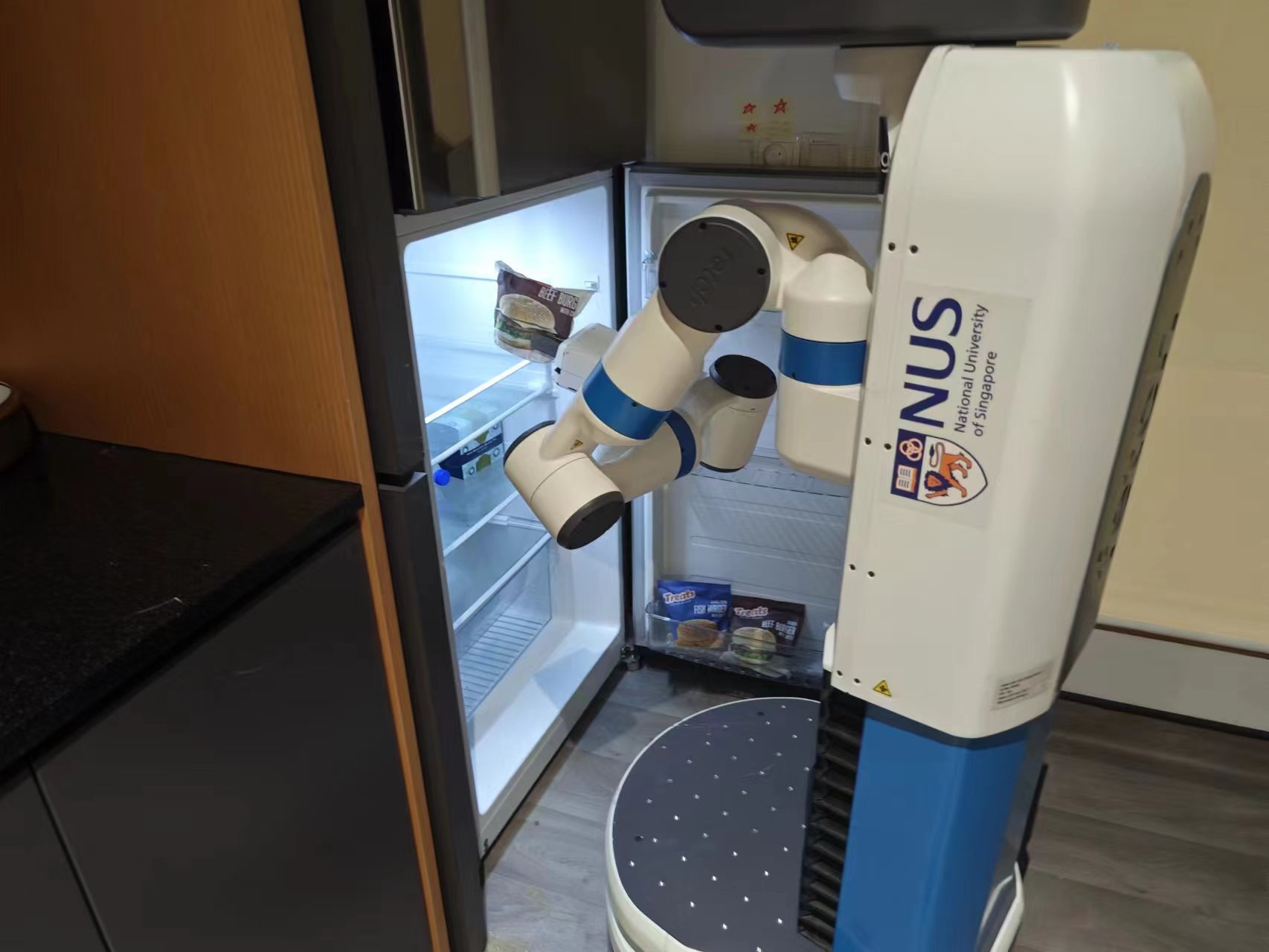}
        \caption{}
        \label{subfig:d}
    \end{subfigure}
        \begin{subfigure}[t]{0.195\linewidth}
        \centering
        \includegraphics[height = 2.6cm]{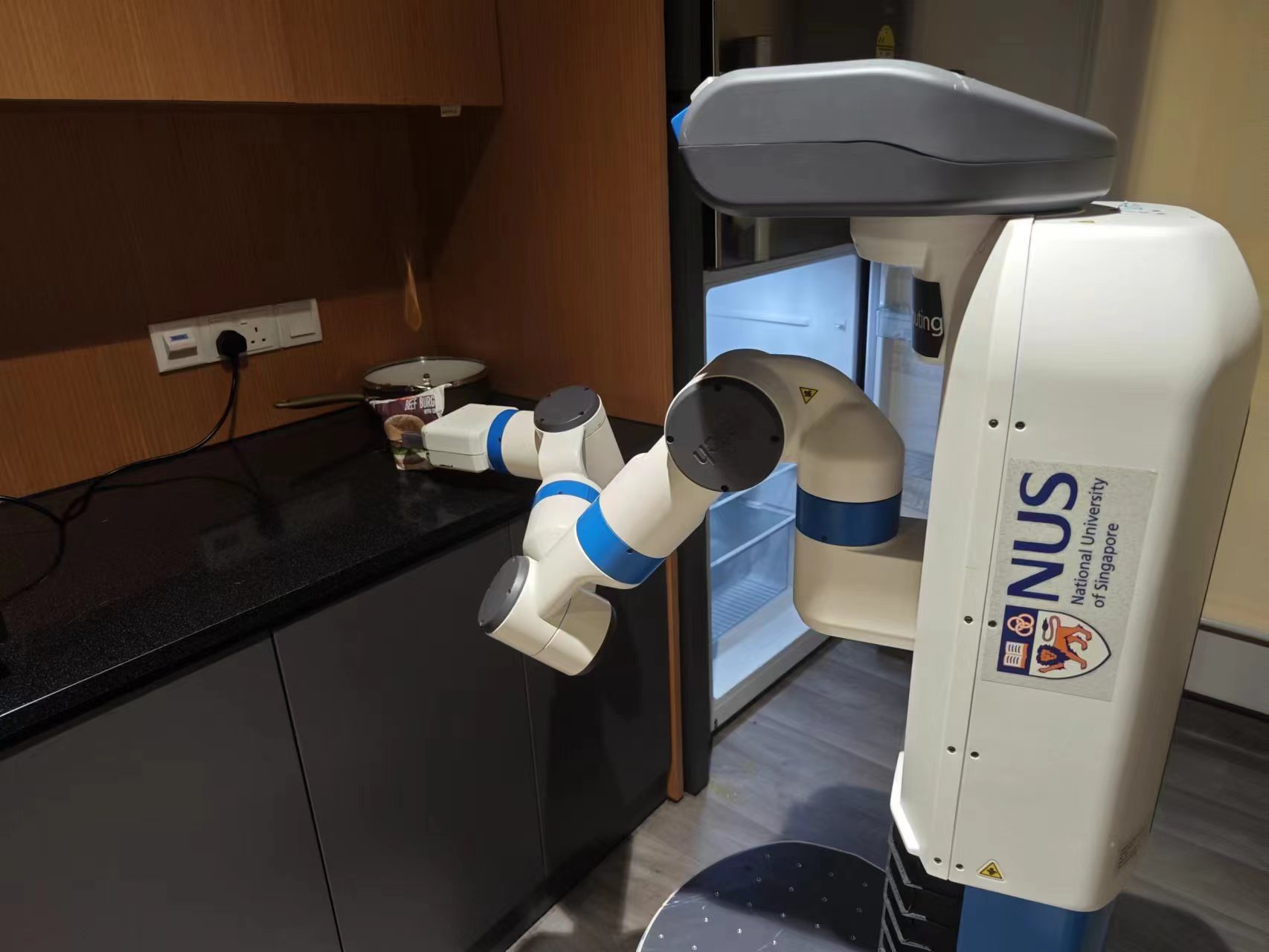}
        \caption{}
        \label{subfig:e}
    \end{subfigure}

        \begin{subfigure}[t]{0.195\linewidth}
        \centering
        \includegraphics[height = 2.6cm]{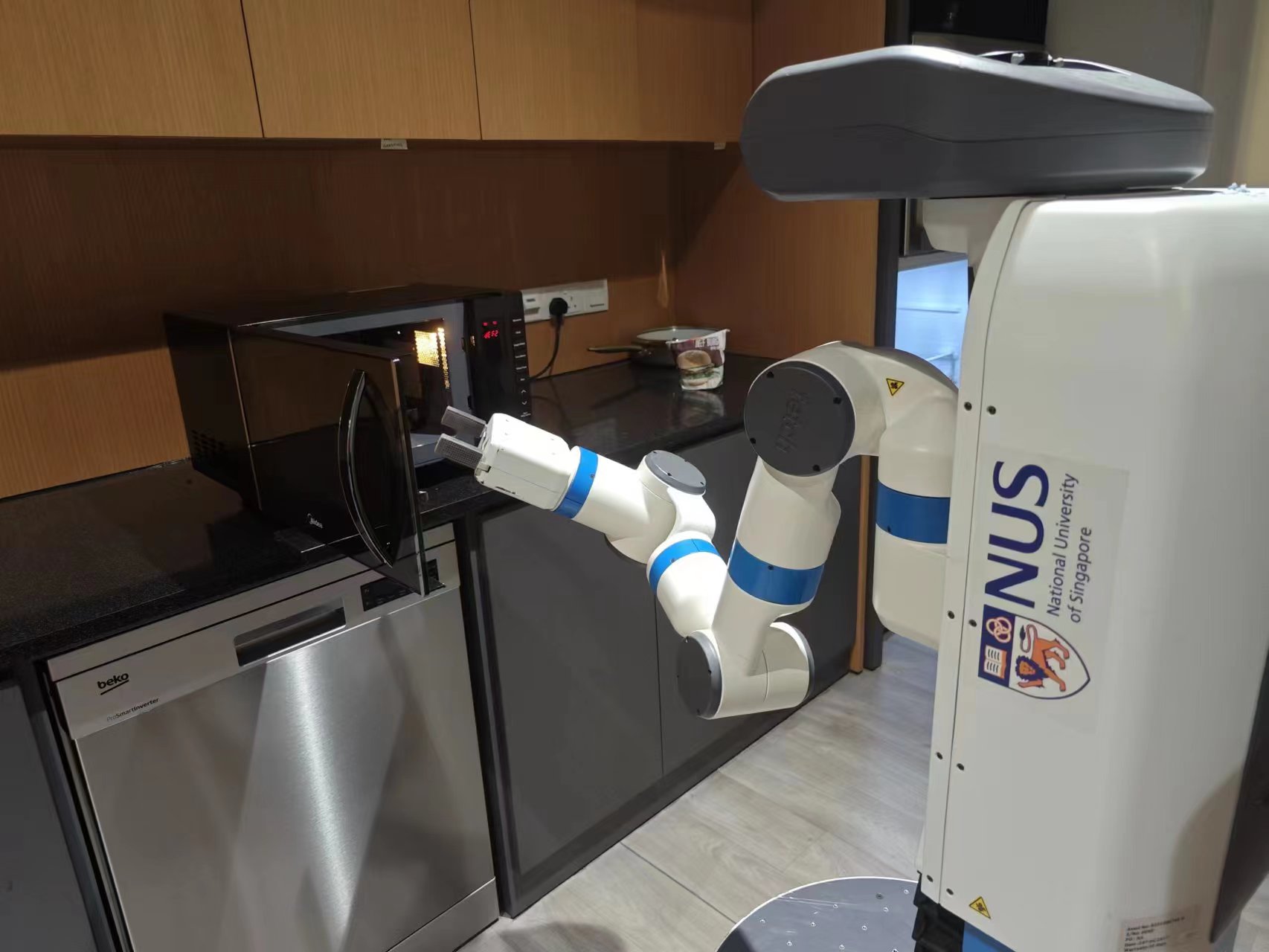}
        \caption{}
        \label{subfig:f}
    \end{subfigure}
    \begin{subfigure}[t]{0.195\linewidth}
        \centering
        \includegraphics[height = 2.6cm]{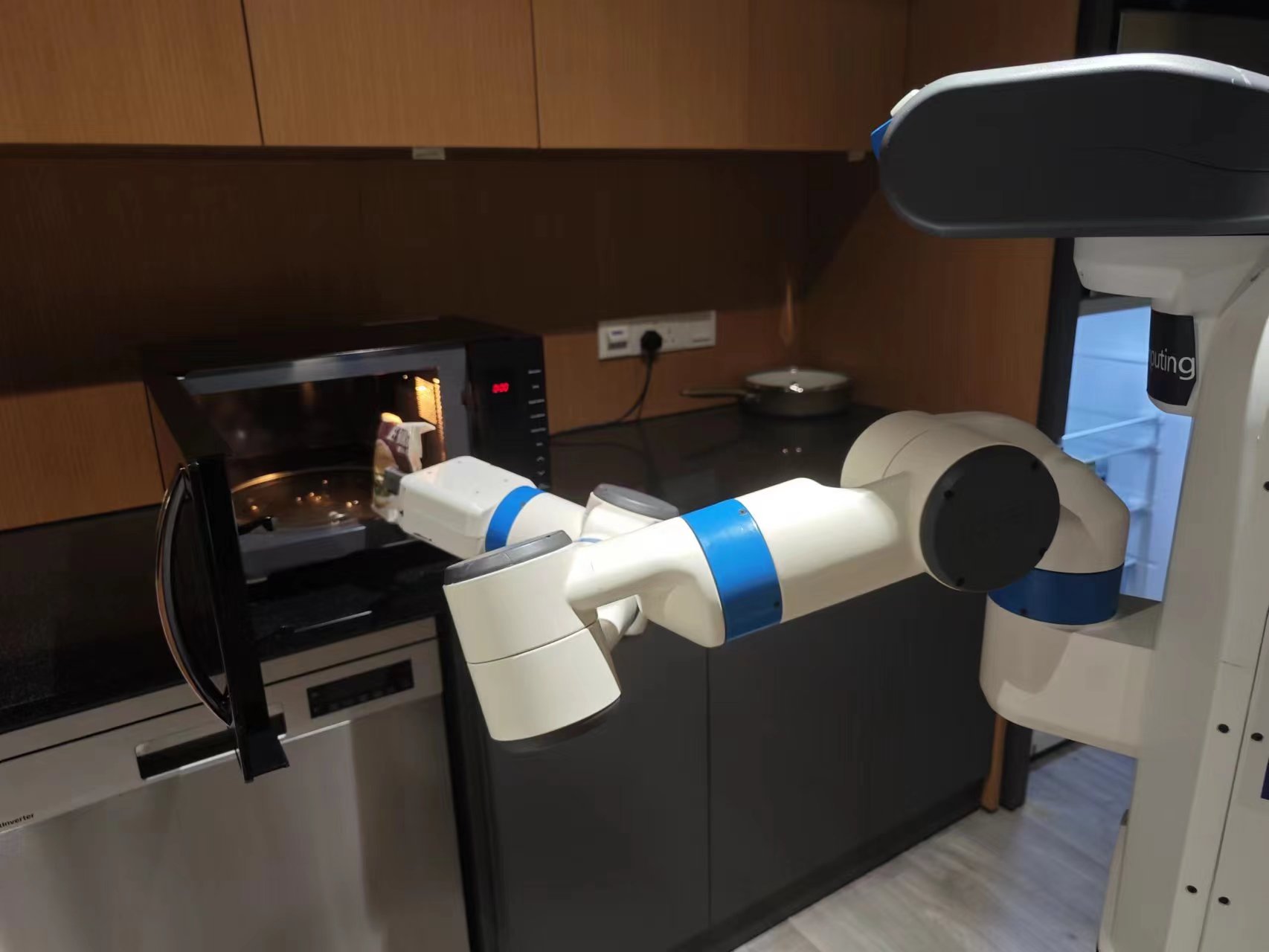}
        \caption{}
        \label{subfig:g}
    \end{subfigure} 
    \centering
    \begin{subfigure}[t]{0.195\linewidth}
        \centering
        \includegraphics[height = 2.6cm]{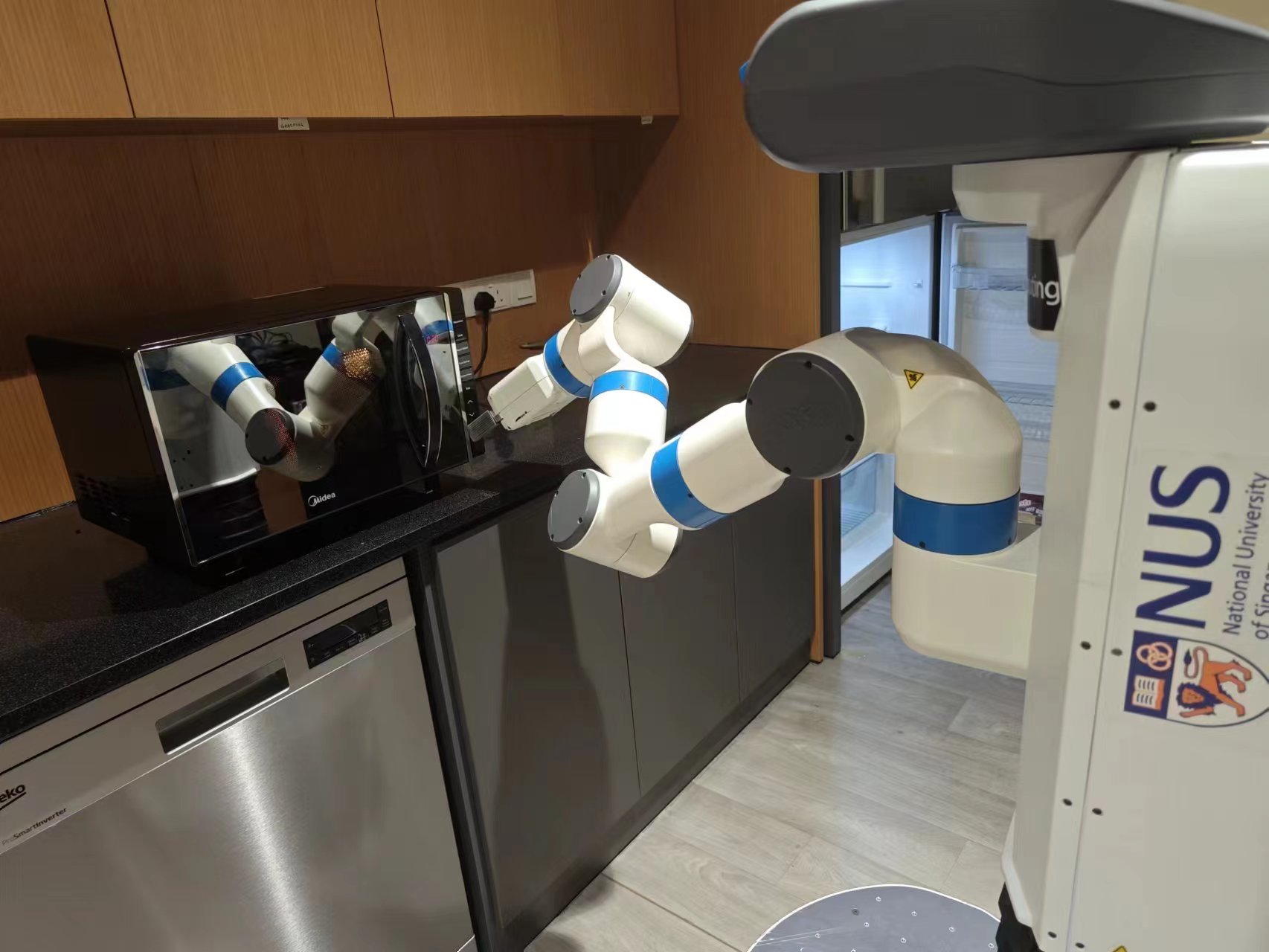}
        \caption{}
        \label{subfig:h}
    \end{subfigure}
    \begin{subfigure}[t]{0.195\linewidth}
        \centering
        \includegraphics[height = 2.6cm]{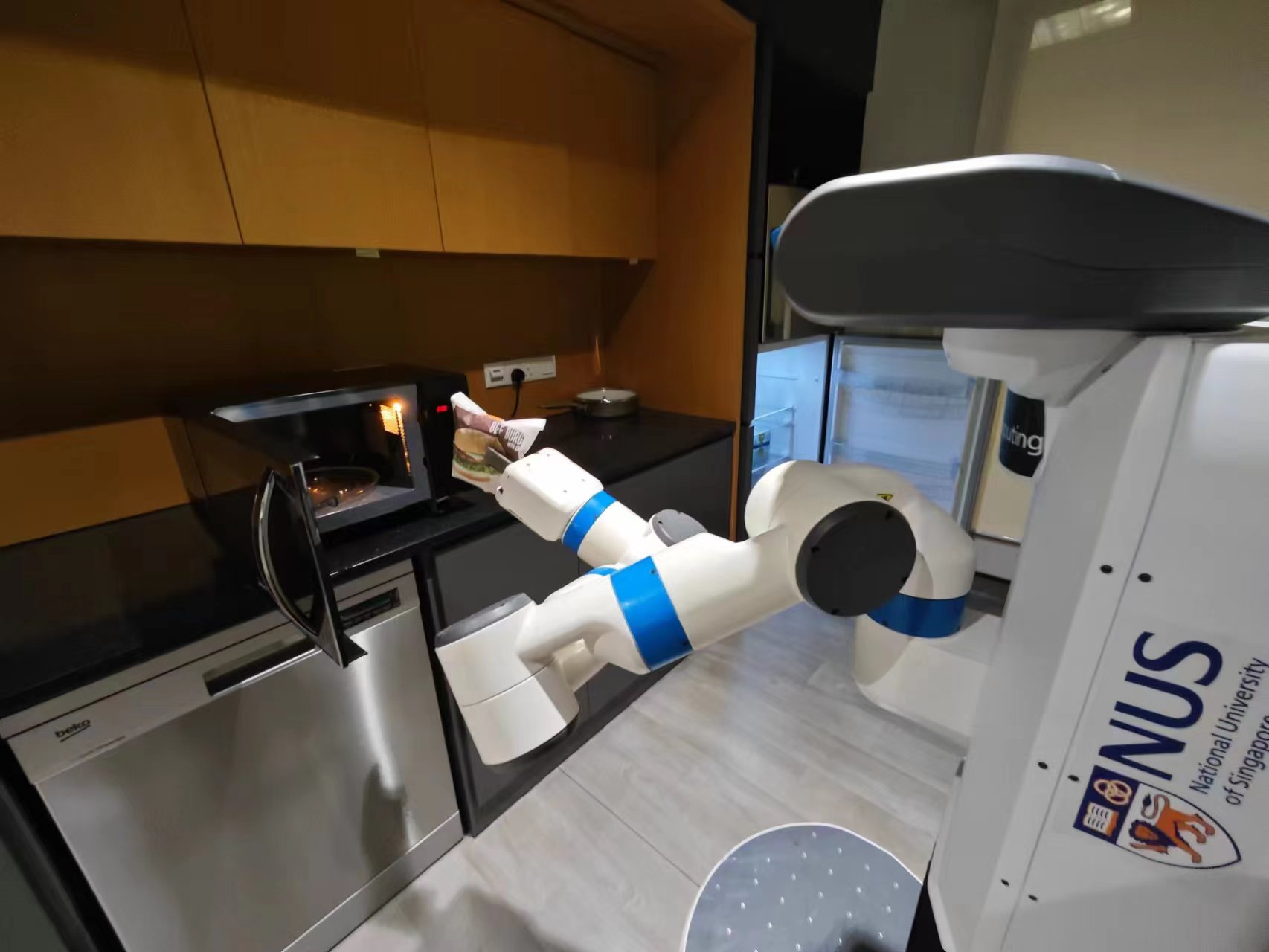}
        \caption{}
        \label{subfig:i}
    \end{subfigure}
        \begin{subfigure}[t]{0.195\linewidth}
        \centering
        \includegraphics[height = 2.6cm]{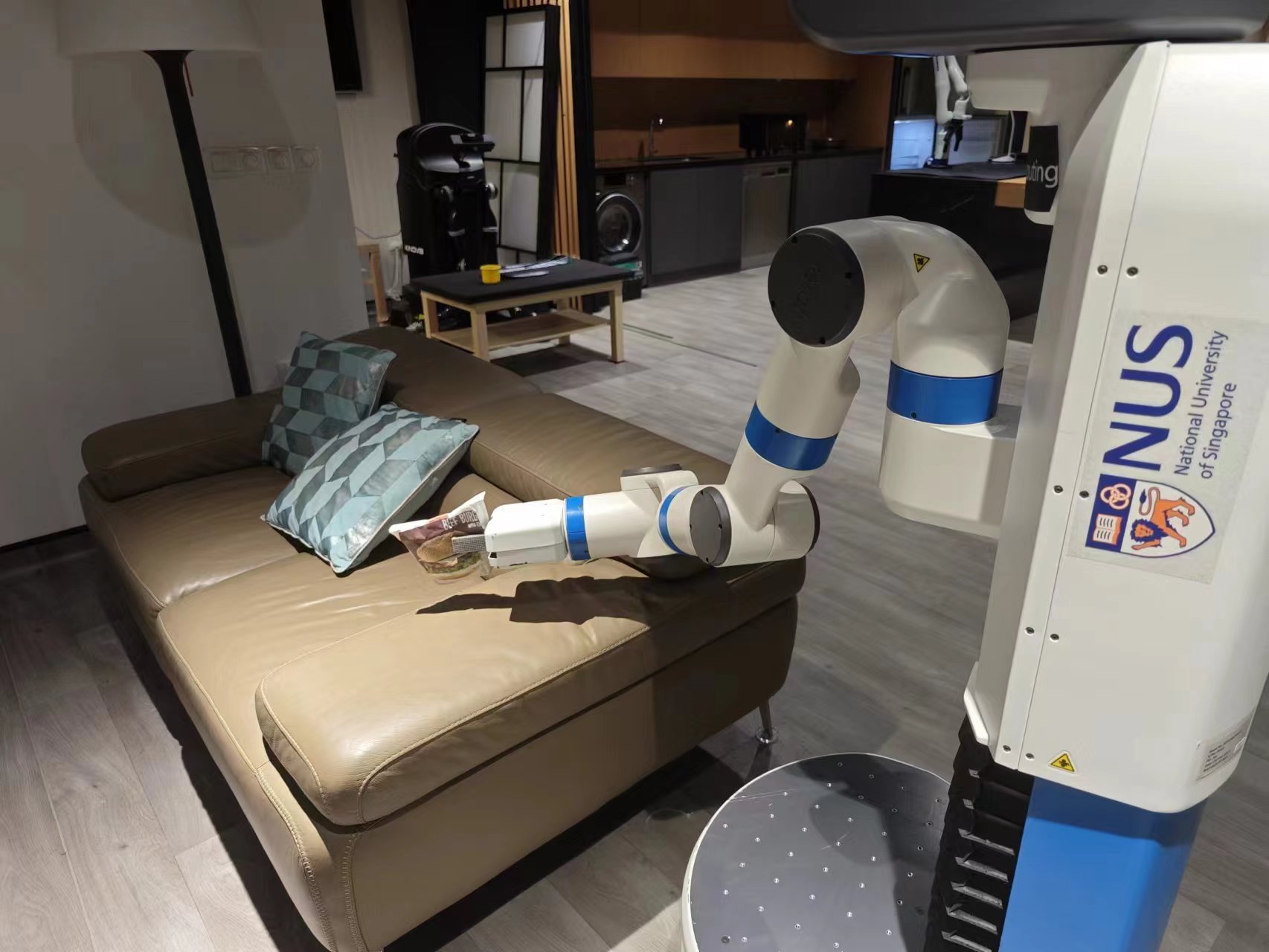}
        \caption{}
        \label{subfig:j}
    \end{subfigure}
    \caption{Snapshot of completing the task \textit{"Heat some food in the fridge and bring it to the sofa."} (a) The robot begins in the living room, after scanning it finds nothing related to the task. (b) The robot navigates to the kitchen and scan the surrounding. (c) The robot opens the fridge to get food. (d) The robot find a hamburger and pick it up. (e) After the unsuccessful attempt to open the microwave, the robot put the food on kitchen counter to clear the hand. (f) The robot successfully opens the microwave. (g) The robot retrieves the food item and places it in the microwave. (h) Following that, the robot switches the microwave on. (i) Once heated, the robot open the microwave and pick up the food. (j) The robot delivers the heated food to the designated sofa.  }
    
    \label{fig:experiment-snapshots}
    \vspace{-0.5cm}
\end{figure*}

The main results of the simulation experiments are shown in Table \ref{table:simulation_result}. We compare the performance of the proposed method with the baselines. The following conclusions can be drawn. 

\textbf{Our method outperforms the baselines on simple and complex tasks.} It achieves a 92.5\% success rate (SR) for simple tasks, surpassing Inner Monologue's 65\% and ProgPrompt's 85\%, indicating the effectiveness of our state representation. For complex tasks, our method achieves an impressive 77.14\% SR, while Inner Monologue and ProgPrompt fail to complete any tasks, which demonstrates the effectiveness of our approach for complex tasks. The failure of Inner Monologue and ProgPrompt to complete long-horizon tasks can be attributed to their lack of state tracking and context understanding.
It's particularly effective on tasks with multiple steps or that require state tracking, improving the success rate on hard tasks from 8.71\% to 77.14\%. Without state tracking, the baseline method fails in complex tasks, whereas our method successfully completes them due to explicit state tracking and reasoning.

We conduct an ablation study to evaluate the impact of individual components in our state representation. The study compares three configurations: one without Retrospective Summary (w/o Summary), another without Object Entries (w/o Objects), and our complete state representation (Our). As shown in Table \ref{table:ablation_study},
\textbf{both unstructured retrospective summary and structured object entries play critical roles in our approach.} In simpler tasks, the performance differences are small. However, in complex tasks, the importance of both components becomes evident. Without retrospective summary, the success rate is only 14.29\%, emphasizing its vital role in handling complex state tracking. The model without object entries achieves a 64.86\% success rate, while our complete state representation reaches 77.14\%, indicating the significance of having a structured state representation for tasks requiring multi-step planning and intricate actions.

\subsection{Real-Robot System Experiments}

\subsubsection{Experimental Settings in Real-world}
To validate the effects of our system in real world with out-of-shelf perception models and low-level controller.
We design our experiments with the Fetch mobile manipulator robot \cite{wise2016fetch} in a typical real home environment consisting of four rooms: kitchen, meeting room, living room and office room. Each room is populated with various objects. 
The system integration was discussed in \ref{subsection:system}. All the computation for decision making, perception and low level control is done on a Linux machine with an NVIDIA RTX 4090 GPU. The perception module is called at the beginning of the task and after the robot has entered a new room or opened a closed container. For safety reasons, when the generated manipulation action command is not in our learned skill library or violates the rules in the real-world, the human monitor will directly deny the execution and the action will return Fail. This process could be achieved automatically by adapting the out-of-distribution (OOD) detection \cite{farid2022task}.

Similar to the simulation experimental setting in \ref{subsubsection:simulation_setting}, we compared the three baselines with our proposed method. 

We evaluate our methods on the real robot with the task: \textit{"heat some food from the fridge and bring it to the sofa"}. This task runs in a challenging open-world setting where there is no food, no fridge, and no heating appliances observed at the start of the task. It is also a long-horizon task, requiring over 20 real robot actions to complete.  For each real executable robot manipulation, we allow 3 retries.  The maximum number of steps is 30. The success is evaluated by the human to check if the goal state in real-world is satisfied.

\begin{table}[t]
\caption{Real-robot experiment results. }

\label{table:real_world}
\centering
\begin{tabular}{@{}llllllll@{}}
\toprule \multicolumn{2}{c}{\multirow{2}{*}{Task}} & \multicolumn{2}{c}{ProgPrompt} & \multicolumn{2}{c}{w/o States} & \multicolumn{2}{c}{Ours} \\  \cmidrule(lr){3-4} \cmidrule(lr){5-6} \cmidrule(lr){7-8}

 &  & SR & AS & SR & AS & SR & AS \\ \midrule
\multicolumn{2}{l}{\begin{tabular}[c]{@{}l@{}}Heat some food from fridge \\ and bring it to the sofa.\end{tabular}} & 0/3 & 30.0 & 0/3 & 30.0 & 3/3 & 21.0 \\ 

\bottomrule
\end{tabular}
\vspace{-0.3cm}
\end{table}

\subsubsection{Real-Robot System Experiment Results}

The results of the real robot system experiments are presented in Table \ref{table:real_world}. All baselines fail to solve this task due to their inability to realize that the robot cannot open the microwave while a food is held by the robot. The lack of explicit state tracking leads to their failure to connect the unsuccessful action \textit{open(microwave)} with the previous action \textit{pick\_up(food)}. In contrast, our method automatically tracks the non-predefined symbolic attribute of the object \textit{\{food: in\_hand\}}, allowing it to capture the actual cause of the failure to open the microwave.
In the experiment shown in Fig. \ref{fig:experiment-snapshots}, the robot effectively completes the extended task. As illustrated in Fig. \ref{subfig:e}, the state representation helps the language model to understand the unsuccessful attempt to open the microwave. Given the explicit state of \textit{\{food: in\_hand\}}, it formulates a new strategy: first put the food on the table, then try to open the microwave and finish the remain steps, thus successfully accomplishing the task.

\section{Conclusion}
In this paper, we presented a novel, dynamic, and expandable representation for LLM doing task planning in open world. Our proposed representation continually expands and updates object attributes based on observable actions and leverages LLM's capability for context understanding and historical action reasoning. Experimental results demonstrate that our proposed representation significantly outperforms baseline methods and enhances the LLM's task planning performance, especially for long-horizon tasks.  Looking forward, a potential improvement could be to include object relations in the state representation. Such information can greatly enhance the LLM's reasoning ability and context understanding, addressing more challenging task-planning problems and further enhancing performance. We believe the proposed representation offers a promising direction for future research in open-world task planning with LLM.

{
    \bibliographystyle{IEEEtran}
    \bibliography{IEEEabrv, bib/bibliography}
}

\section*{Appendix}
\definecolor{darkgreen}{rgb}{0.0, 0.7, 0.0}
\definecolor{darkblue}{rgb}{0.0, 0.0, 0.7}
\lstdefinestyle{mystyle}{
    moredelim=**[is][\color{darkblue}]{`}{`},
    moredelim=**[is][\color{darkgreen}]{^}{^},
    moredelim=**[is][\color{red}]{\#}{\#}}
\lstset{style=mystyle}

\subsection{Detailed Prompt}
\label{appendix:prompt}
We provided detailed prompts from each component of our approach: LLM as Attention, LLM as State Estimator, and LLM as Policy. The \textcolor{darkblue}{blue} color refers to dynamically injected content, the \textcolor{darkgreen}{green} refers to LLM generated content, and black color refers to fixed content.

\begin{lstlisting}[caption=Example Prompt of LLM as Attention]
user: 
******** current state start ********
`lightswitch1: []
bedroom1: []`

******** current state end ********

******** current observation start ********
`Object List: ['towelrack1 IN bathroom1', 'towelrack2 IN bathroom1', 'towelrack3 IN bathroom1', 'towelrack4 IN bathroom1', 'faucet1 ON bathroomcounter1', 'towel1 ON bathroomcounter1', 'towel2 ON bathroomcounter1', 'towel3 ON towel1', 'hairproduct1 IN bathroom1', 'hairproduct2 IN bathroom1', 'facecream1 ON bathroomcounter1', 'plate1 ON wallshelf1', 'toothpaste1 ON bathroomcounter1', 'toothbrush1 ON bathroomcounter1', 'barsoap1 ON bathroomcounter1', 'towel4 IN bathroom1', 'towel5 IN bathroom1', 'candle1 ON wallshelf1', 'lightswitch1 IN bathroom1', 'washingmachine1 IN bathroom1', 'tablelamp1 ON nightstand1', 'tablelamp2 ON nightstand2', 'chair1 IN bedroom1', 'chair2 IN bedroom1', 'mouse1 ON desk1', 'keyboard1 ON desk1', 'lightswitch2 IN bedroom1', 'computer1 IN bedroom1', 'radio1 IN bookshelf1', 'cellphone1 ON coffeetable1', 'book1 ON bookshelf1', 'book2 IN bookshelf1', 'box1 IN bookshelf1', 'plate2 ON desk1', 'mug1 ON desk1', 'cupcake1 ON desk1', 'cupcake2 ON desk1', 'plate3 ON coffeetable1', 'slippers1 IN bedroom1', 'slippers2 IN bedroom1', 'cellphone2 ON coffeetable1', 'folder1 ON bookshelf1', 'folder2 ON bookshelf1']
Container List: ['toilet1 IN bathroom1', 'bathroomcabinet1 IN bathroom1', 'washingmachine1 IN bathroom1', 'nightstand1 IN bedroom1', 'nightstand2 IN bedroom1', 'closet1 IN bedroom1', 'closet2 IN bedroom1']
Surface List: ['towelrack1 IN bathroom1', 'towelrack2 IN bathroom1', 'towelrack3 IN bathroom1', 'towelrack4 IN bathroom1', 'bathroomcabinet1 IN bathroom1', 'bathroomcounter1 IN bathroom1', 'plate1 ON wallshelf1', 'nightstand1 IN bedroom1', 'bookshelf1 IN bedroom1', 'nightstand2 IN bedroom1', 'chair1 IN bedroom1', 'desk1 IN bedroom1', 'bed1 IN bedroom1', 'chair2 IN bedroom1', 'coffeetable1 IN bedroom1', 'mousemat1 ON desk1', 'radio1 IN bookshelf1', 'plate2 ON desk1', 'plate3 ON coffeetable1']
Room List: ['bathroom1', 'bedroom1', 'kitchen1', 'livingroom1']
Current Holding: []
You are closed to: ['plate3', 'cellphone1', 'coffeetable1', 'cellphone2']
Current Room: bedroom1`
******** current observation end ********

Task goal: `switch off all the lights in the house`

Given the current observation and task goal, please add related objects from observations:
add_related_objects(object)

for example:
Task goal: peel an apple
add_related_objects("apple1")
add_related_objects("apple2")
add_related_objects("knife1")
            
LLM response: 
^add_related_objects("lightswitch1")
add_related_objects("lightswitch2")^

\end{lstlisting}

\begin{lstlisting}[caption=Example Prompt of LLM as State Estimator]
user: 
******** current state start ********
`lightswitch1: []
bedroom1: []
lightswitch2: []`

******** current state end ********

******** robot actions start ********
`["['move', 'lightswitch1'](Success)", "['switchoff', 'lightswitch1'](Success)", "['move', 'bedroom1'](Success)", "['switchoff', 'lightswitch1'](Fail)"]`
******** robot actions end ********

`Room List: ['bathroom1', 'bedroom1', 'kitchen1', 'livingroom1']
Current Holding: []
You are closed to: ['plate3', 'cellphone1', 'coffeetable1', 'cellphone2']
Current Room: bedroom1`

Important! 
You may need to search different places and containers for relevant items.
You can only interact (e.g. pickup, open, switch on) with objects that are closed to you.
You only have two hands, when you hold two items, you cannot pickup/open/switch on other items.
    

Given the current state and robot actions, please update the state by calling the following function:
update_state(object, updated_status)
update_reasoning("reasoning process")

for example:
update_reasoning("fill in the reasoning here especially for possible failure cases")
update_state("apple", "on_table | in_hand")
update_state("table", "in kitchen | not_graspable | not_movable")
        
LLM response: 
^update_reasoning("The robot moved to lightswitch1 and successfully switched it off. Then it moved to bedroom1. It tried to switch off lightswitch1 again but failed because it's not in the same location with lightswitch1.")
update_state("lightswitch1", "off")
update_state("bedroom1", "robot_inside")^


\end{lstlisting}

\begin{lstlisting}[caption=Example Prompt of LLM as Policy,
escapeinside={(*}{*)}]
user: 
******** instruction start ********

At any point, you only have access to the following functions: 
    move(a), pickup(a), placein(a,b), placeon(a,b), open(a), close(a), switchon(a), switchoff(a), wait()
    
******** instruction end ********


********  response template start ********
Low-level Action Plan:
1. move(fridge) (example only)
2. open(fridge) (example only)
3. move(cup1) (example only)
4. pickup(cup1) (example only)
5. move(cupboard1) (example only)
6. placein(cup1, cupboard1) (example only)
******** response template end ********

******** history actions start ********
`["['move', 'lightswitch1'](Success)", "['switchoff', 'lightswitch1'](Success)", "['move', 'bedroom1'](Success)", "['switchoff', 'lightswitch1'](Fail)"]`
******** history actions end ********

******** current observation start ********
`Object List: ['towelrack1 IN bathroom1', 'towelrack2 IN bathroom1', 'towelrack3 IN bathroom1', 'towelrack4 IN bathroom1', 'faucet1 ON bathroomcounter1', 'towel1 ON bathroomcounter1', 'towel2 ON bathroomcounter1', 'towel3 ON towel1', 'hairproduct1 IN bathroom1', 'hairproduct2 IN bathroom1', 'facecream1 ON bathroomcounter1', 'plate1 ON wallshelf1', 'toothpaste1 ON bathroomcounter1', 'toothbrush1 ON bathroomcounter1', 'barsoap1 ON bathroomcounter1', 'towel4 IN bathroom1', 'towel5 IN bathroom1', 'candle1 ON wallshelf1', 'lightswitch1 IN bathroom1', 'washingmachine1 IN bathroom1', 'tablelamp1 ON nightstand1', 'tablelamp2 ON nightstand2', 'chair1 IN bedroom1', 'chair2 IN bedroom1', 'mouse1 ON desk1', 'keyboard1 ON desk1', 'lightswitch2 IN bedroom1', 'computer1 IN bedroom1', 'radio1 IN bookshelf1', 'cellphone1 ON coffeetable1', 'book1 ON bookshelf1', 'book2 IN bookshelf1', 'box1 IN bookshelf1', 'plate2 ON desk1', 'mug1 ON desk1', 'cupcake1 ON desk1', 'cupcake2 ON desk1', 'plate3 ON coffeetable1', 'slippers1 IN bedroom1', 'slippers2 IN bedroom1', 'cellphone2 ON coffeetable1', 'folder1 ON bookshelf1', 'folder2 ON bookshelf1']
Container List: ['toilet1 IN bathroom1', 'bathroomcabinet1 IN bathroom1', 'washingmachine1 IN bathroom1', 'nightstand1 IN bedroom1', 'nightstand2 IN bedroom1', 'closet1 IN bedroom1', 'closet2 IN bedroom1']
Surface List: ['towelrack1 IN bathroom1', 'towelrack2 IN bathroom1', 'towelrack3 IN bathroom1', 'towelrack4 IN bathroom1', 'bathroomcabinet1 IN bathroom1', 'bathroomcounter1 IN bathroom1', 'plate1 ON wallshelf1', 'nightstand1 IN bedroom1', 'bookshelf1 IN bedroom1', 'nightstand2 IN bedroom1', 'chair1 IN bedroom1', 'desk1 IN bedroom1', 'bed1 IN bedroom1', 'chair2 IN bedroom1', 'coffeetable1 IN bedroom1', 'mousemat1 ON desk1', 'radio1 IN bookshelf1', 'plate2 ON desk1', 'plate3 ON coffeetable1']
Room List: ['bathroom1', 'bedroom1', 'kitchen1', 'livingroom1']
Current Holding: []
You are closed to: ['plate3', 'cellphone1', 'coffeetable1', 'cellphone2']
Current Room: bedroom1`
******** current observation end ********

******** current state start ********
`lightswitch1: off
bedroom1: robot_inside
lightswitch2: []


Reasoning: The robot moved to lightswitch1 and successfully switched it off. Then it moved to bedroom1. It tried to switch off lightswitch1 again but failed because it's not in the same location with lightswitch1.`
******** current state end ********


Important! 
You may need to search different places and containers for relevant items.
You can only interact (e.g. pickup, open, switch on) with objects that are closed to you.
You only have two hands, when you hold two items, you cannot pickup/open/switch on other items.
    

Task goal: `switch off all the lights in the house`
- Generate a plan based on the current state.
- Make the plan concise and short, less than `20` steps.
- Be careful about the state information, consider the additional notes in the state.
- You may need to search different places and containers for relevant items.

LLM response: 
^Low-level Action Plan:
1. move(lightswitch2)
2. switchoff(lightswitch2)
3. move(kitchen1)
4. move(lightswitch3) (assuming there is a lightswitch3 in the kitchen)
5. switchoff(lightswitch3)
6. move(livingroom1)
7. move(lightswitch4) (assuming there is a lightswitch4 in the living room)
8. switchoff(lightswitch4)
9. move(bathroom1)
10. move(lightswitch1)
11. switchoff(lightswitch1)^


\end{lstlisting}

\subsection{Unstructured Summary Examples}
\label{sec:summary_examples}
In this section, we provide examples of how the unstructured summary helps with better object attribute prediction.

\subsubsection{Without Summary} Below is an example of not using the unstructured summary.

\begin{lstlisting}[caption=Example of Unstructured Summary Helps with Better Object Attributes prediction.]
******** history actions start ********
["['move', 'kitchen1'](Success)", "['move', 'kitchentable1'](Success)", "['pickup', 'item1'](Fail)", "['move', 'kitchentable1'](Success)", "['pickup', 'cutleryknife2'](Success)", "['move', 'kitchencabinet1'](Success)", "#['placein', 'cutleryknife2', 'kitchencabinet1'](Fail)#"]
******** history actions end ********

******** current observation start ********
Object List: ['towelrack1 IN bathroom1', 'towelrack2 IN bathroom1', 'towelrack3 IN bathroom1', 'towelrack4 IN bathroom1', 'faucet1 ON bathroomcounter1', 'towel1 ON bathroomcounter1',
... // contents omitted for clarity
******** current observation end ********

******** current state start ********
kitchen1: []
kitchentable1: []
cutleryknife2: in_hand
#kitchencabinet1: []#
kitchencabinet2: []
******** current state end ********

\end{lstlisting}

\subsubsection{With Summary} Below is an example of using the unstructured summary.

\begin{lstlisting}[caption=Example of Unstructured Summary Helps with Better Object Attributes prediction.]
******** history actions start ********
["['move', 'kitchen1'](Success)", "['move', 'kitchentable1'](Success)", "['pickup', 'item1'](Fail)", "['move', 'kitchentable1'](Success)", "['pickup', 'cutleryknife2'](Success)", "['move', 'kitchencabinet1'](Success)", "#['placein', 'cutleryknife2', 'kitchencabinet1'](Fail)#"]
******** history actions end ********

******** current observation start ********
Object List: ['towelrack1 IN bathroom1', 'towelrack2 IN bathroom1', 'towelrack3 IN bathroom1', 'towelrack4 IN bathroom1', 'faucet1 ON bathroomcounter1', 'towel1 ON bathroomcounter1',
... // contents omitted for clarity
******** current observation end ********

******** current state start ********
Reasoning: The robot successfully moved to the kitchen and the kitchen table. It failed to pick up 'item1' because 'item1' is not specified in the current state. The robot then moved to 'cutleryknife2' and successfully picked it up. It then moved to 'kitchencabinet1' and tried to place 'cutleryknife2' in 'kitchencabinet1', but failed. This could be because 'kitchencabinet1' is not open.

kitchen1: []
kitchentable1: []
cutleryknife2: in_hand
#kitchencabinet1: closed | in_kitchen#
kitchencabinet2: []
******** current state end ********

\end{lstlisting}

\end{document}